\documentclass{svmult}

\usepackage{type1cm} 
\usepackage{verbatim}
\usepackage{amsmath}
\usepackage{bm}
\usepackage{makeidx}         
\usepackage{graphicx}        
\usepackage{multicol}        
\usepackage[bottom]{footmisc}

\usepackage{newtxtext}       %
\usepackage{newtxmath}       
\usepackage{float}

\makeindex             

\begin{document}

\title*{Towards Adaptive Fusion of Multimodal Deep Networks for Human Action Recognition}
\author{Novanto Yudistira}
\institute{Novanto Yudistira \at Departemen Teknik Informatika, Fakultas Ilmu Komputer, Universitas Brawijaya, Indonesia \email{yudistira@ub.ac.id}}
%
%
\maketitle

\abstract{This study introduces a pioneering methodology for human action recognition by harnessing deep neural network techniques and adaptive fusion strategies across multiple modalities, including RGB, optical flows, audio, and depth information. Employing gating mechanisms for multimodal fusion, we aim to surpass limitations inherent in traditional unimodal recognition methods while exploring novel possibilities for diverse applications. Through an exhaustive investigation of gating mechanisms and adaptive weighting-based fusion architectures, our methodology enables the selective integration of relevant information from various modalities, thereby bolstering both accuracy and robustness in action recognition tasks. We meticulously examine various gated fusion strategies to pinpoint the most effective approach for multimodal action recognition, showcasing its superiority over conventional unimodal methods. Gating mechanisms facilitate the extraction of pivotal features, resulting in a more holistic representation of actions and substantial enhancements in recognition performance. Our evaluations across human action recognition, violence action detection, and multiple self-supervised learning tasks on benchmark datasets demonstrate promising advancements in accuracy. The significance of this research lies in its potential to revolutionize action recognition systems across diverse fields. The fusion of multimodal information promises sophisticated applications in surveillance and human-computer interaction, especially in contexts related to active assisted living.}
\keywords{Multimodal fusion; Deep neural networks; Human action recognition; Gating mechanisms; Adaptive fusion strategies}

\section{Introduction}
\label{sec:1}
Our society is experiencing a significant demographic shift due to the aging population, which is why there is an increasing demand for cutting-edge technologies that support active and assisted living. This demographic trend highlights how critical it is to address the unique challenges faced by the elderly, those with disabilities, and those in need of assisted living services, according to Rashidi and Mihailidis \cite{rashidi2012survey}. Enhancing the quality of life of these populations through the combination of deep learning techniques, artificial intelligence (AI), and multimodal data fusion seems to be a promising approach \cite{nweke2019data}\cite{qiu2022multi}.

The work of Nweke et al. and Qiu et al. provides direction in the rapidly changing field of technology, showing the way toward a time when adaptive fusion methods will be essential to the independence and well-being of people residing in assisted and active care \cite{nweke2019data}\cite{qiu2022multi}. Their contributions not only advance our current knowledge but also provide the groundwork for future research, which is a major step in the direction of designing responsive, caring, and technologically advanced living spaces.

The interdisciplinary nature of this exploration delves into the intricate dance between artificial intelligence and the fusion of multimodal data. It seeks to decipher how these technologies can adapt to the unique needs of individuals, creating a symbiotic relationship between the technology and the user \cite{zhang2020multimodal}. Especially for video based recognition, as we navigate the complexities of designing intelligent environments, the pivotal role of adaptive fusion becomes increasingly apparent, facilitating a seamless integration of technology into the fabric of daily life for seniors and individuals with specific care requirements \cite{shivappa2010audiovisual}\cite{watanabe2023technology}\cite{gray2016supporting}. Active and assisted living environments demand adaptability and customization unlike any other, considering the unique needs and preferences of individuals. Conventional single-modal methods frequently fail to record the rich data necessary for the best possible care. Using deep neural networks to combine multi - modal sources of data as vision, audio, and sensor data—can be customized for for every person, offering a novel paradigm for assisting with day-to-day living \cite{bayoudh2021survey}.

The task of video classification has become interesting in computer vision and pattern recognition due to dynamic scenes and objects with spatial or temporal variations \cite{simonyan2014two}\cite{yudistira2017gated}. This simply creates suitable handcrafted features challenging. Convolutional neural networks (CNNs) have revolutionized feature learning by considering multiple factors, such as neighboring pixels and shapes. While deep neural networks (CNNs) are effective at categorizing videos, a gating mechanism is still needed to balance the relative importance of the different modalities. For example, the gating network should prioritize spatial cues and vice versa if spatial cues are more salient than temporal ones. Deep neural networks are excellent at processing complex data and extracting meaningful information, which makes them useful in the context of assisted and active living. The investigation centers on adaptive fusion, which goes beyond simple data integration to emphasize the significance of tailoring to residents' particular requirements and preferences, whether it be for tracking health indicators, identifying everyday activities, or identifying falls.

Large Language Models (LLMs), originally designed for natural language processing tasks, are now rapidly expanding into the multimodal domain. Traditional LLMs excel at textual reasoning but lack perceptual capabilities, while large vision models (LVMs) possess strong visual feature extraction but limited symbolic reasoning \cite{yin2023multimodal}. The combination of these complementary strengths has led to the emergence of \emph{Multimodal Large Language Models} (MLLMs), enabling unified reasoning over diverse data modalities \cite{han2024onellm}. This paradigm has produced state-of-the-art systems such as Flamingo and GPT-4 Vision, which perform robust image, video, and real-world contextual understanding across modalities \cite{alayrac2022flamingo, openai2023gpt4}.

In human action recognition (HAR), LLM integration is gaining momentum as a means to incorporate contextual reasoning and prior knowledge into multimodal systems. Recent studies demonstrate that LLMs can directly serve as action recognizers: one framework encodes skeleton-motion sequences into descriptive ``action sentences'' and feeds them to a pretrained LLM (without fine-tuning), relying on the model’s implicit knowledge of human behavior \cite{qu2024llmar}. Meanwhile, trends in adaptive fusion increasingly position LLMs as the central component of multimodal integration—features from sensors, audio, or vision are projected into a shared textual space, allowing unified LLM-based reasoning \cite{li2023blip2}. Other architectures introduce modality-specific adapters inside LLMs to achieve flexible, end-to-end multimodal alignment \cite{bai2023qwenvl}. Such approaches promise richer and more semantically grounded representations of human actions than traditional unimodal or static fusion schemes.

The difficulties of managing diverse data sources and guaranteeing security and privacy are explored throughout the investigation. Case studies and real-world applications show how adaptive multimodal fusion enhances people's lives by fostering self-reliance, well-being, and a higher standard of living. This section begins an exploration of adaptive fusion techniques in the complex field of assisted and active living. These methods are the foundation of responsive, intelligent systems made to accommodate people's various needs and complex preferences in these kinds of living spaces. Nweke et al. and Qiu et al. have made priceless contributions to the field of data fusion, highlighting its significance in developing adaptive systems that surpass traditional methods. \cite{qiu2022multi}\cite{nweke2019data}.

Adaptive fusion is a technique that actively combines artificial intelligence and deep learning to manage the convergence of multiple sensory inputs \cite{\cite{qiu2022multi}}. This convergence is necessary to build systems that can perceive and process a broad variety of inputs and intelligently adjust to the intricacies of day-to-day living in assisted and active living environments. Because of the depth of understanding gained through adaptive fusion techniques, these systems can identify subtle cues, providing residents with a personalized and responsive experience.

This chapter provides insights into current technological advancements and explores possibilities for the future and their implications for society. The conversation concludes that a critical first step toward creating a more compassionate world is an adaptive fusion of multimodal deep networks. In addition to being a significant technological development, it is also an essential tool for advancing the well-being and independence of those residing in assisted and active living.

\section{Multimodal Data Sources}
\label{sec:2}
Capturing a complex spectrum of data about the environment and the residents themselves is essential in the dynamic living support domain, where the goal is to provide people with comprehensive assistance \cite{gaohuman}. This is the point at which using multimodal data sources becomes essential. These sources encompass a variety of sensory inputs, each of which adds a distinct perspective to the overall comprehension of the situation. This section looks at the main multimodal data sources that are frequently used in assisted and active living settings and their significance for creating comprehensive, context-aware systems. The input consists of camera-based vision along with text, audio, and sensor data. This makes it difficult to comprehend how these modalities can work together to provide additional information.

Systems for assisted diverse and complex active living are becoming more and more dependent on a range of sensory inputs to create environments at the same time that technology is advancing. Apart from traditional camera-based vision, sensor data provides instantaneous feedback on actions and movements, text inputs reveal patterns of interactions and communication, and audio inputs offer auditory context. The amalgamation of these multimodal sources facilitates a more profound comprehension of the residents' requirements, inclinations, and everyday regimens. As these systems develop, the challenge is to enable the smooth integration of various modalities, making sure that each maintains security and privacy while making a significant contribution to the overall context. Examining these multimodal nuances highlights how crucial it is to adapt to the varied ways that people experience these environments and enhances the functionality of assisted and active living technologies.

\subsection{Vision-Based Data}
\label{subsec:2}
In this context, visual data—which includes a variety of visual information taken by cameras and imaging devices—is a basic part of multimodal data. Applications like facial recognition, object detection, and scene analysis are made possible by the real-time visual data that vision sensors provide \cite{bux2017vision}. These devices are useful for tracking whereabouts, keeping an eye on residents' activities, and guaranteeing their safety. In fall detection and security applications, vision-based data is essential, especially when it comes to comprehending the spatial dynamics of the living space \cite{harrou2017vision} \cite{basavaraj2017vision}. Modalities such as RGB (Red, Green, Blue), optical flows \cite{beauchemin1995computation}, depth \cite{torralba2002depth}, and skeleton modality \cite{duan2022revisiting} can be strategically employed. RGB data provides color information, offering a nuanced perspective, while optical flows capture motion patterns, revealing insights into dynamic activities. Depth data adds a three-dimensional dimension for a more comprehensive environmental understanding. The skeleton modality, focusing on structural aspects of the human body, can extract specific information tailored to application requirements \cite{duan2022revisiting}.

\begin{figure}[H]
\centering
\includegraphics[scale=.2]{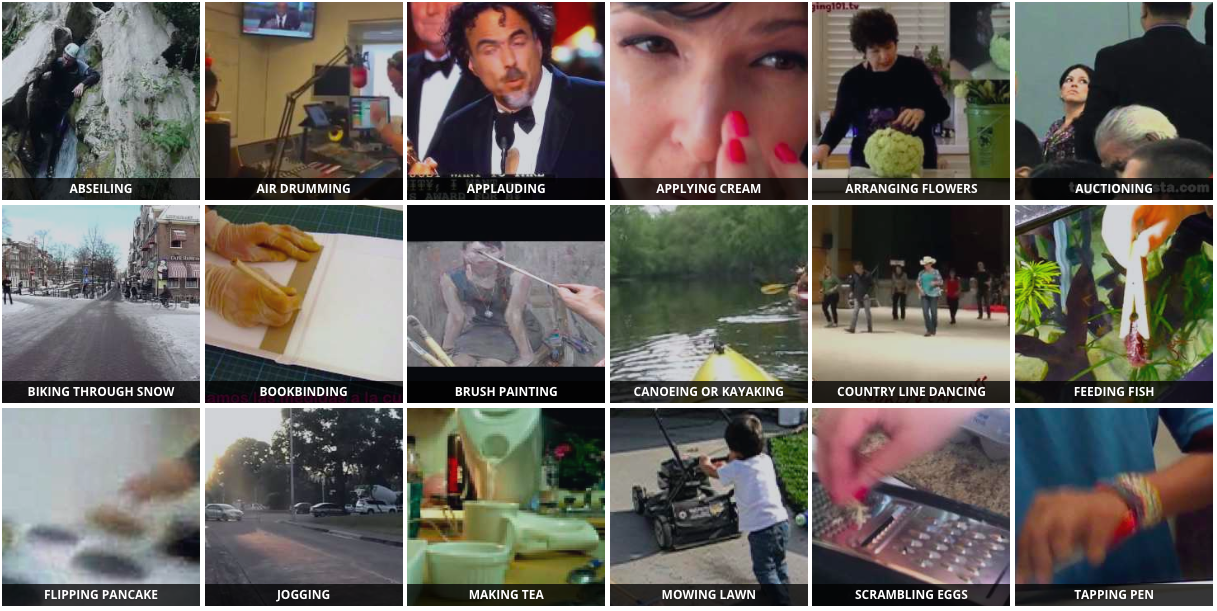}
%
%
\centering
\caption{Several RGB modalities representing human actions \cite{kuehne2011hmdb}}
\label{fig:rgb}       
\end{figure}

Several RGB modalities (Fig. \ref{fig:rgb}) may encompass different approaches or variations in utilizing RGB data for human action recognition. For instance, it could refer to distinct processing techniques or feature extraction methods applied to RGB data to highlight specific aspects of human actions. These modalities might include variations in the representation of color, texture, or spatial information within the RGB data, enabling a more nuanced analysis of human actions.

\begin{figure}[h]
\centering
\includegraphics[scale=.4]{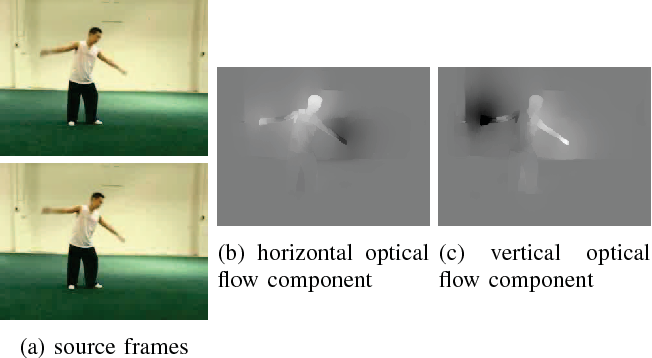}
%
%
\centering
\caption{Optical Flows of Human Action}
\label{fig:of}       
\end{figure}

Optical flows (Fig.\ref{fig:of}) can be expressed as a vector field, where each vector has both horizontal and vertical components. The horizontal component represents the motion of objects or pixels along the horizontal axis, while the vertical component represents the motion along the vertical axis. This decomposition allows for a more detailed understanding of the direction and magnitude of motion in different spatial dimensions.

\begin{figure}[h]
\centering
\includegraphics[scale=.6]{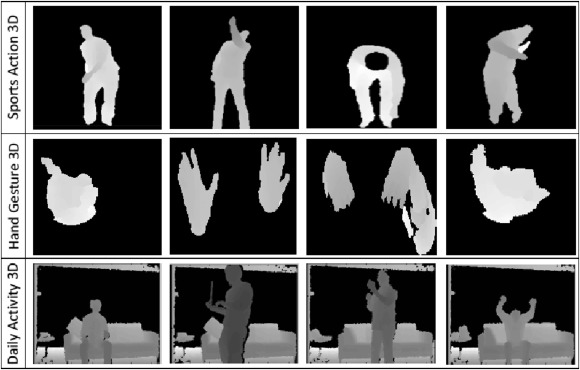}
%
%
\centering
\caption{Depth modality as representation of sports actions, hand gestures, and daily activity \cite{ali2018depth}}
\label{fig:depth}       
\end{figure}

Depth modalities provide a more thorough understanding of human movements and interactions with the environment in the context of daily activities. It helps with tracking object manipulation, identifying body postures, and comprehending the spatial context of daily activities. When compared to traditional visual data, depth sensors provide a richer representation of daily activities because they are able to capture subtle movements and gestures.

The significance of three-dimensional information in comprehending and interpreting human actions is highlighted by the application of depth modalities (Fig. \ref{fig:depth}) as a representation for hand gestures, sports actions, and everyday activities. By improving the depth of analysis, this modality offers insightful information about spatial relationships and dynamics that may be difficult to discern from conventional two-dimensional visual data.

\begin{figure}[h]
\centering
\includegraphics[scale=1.]{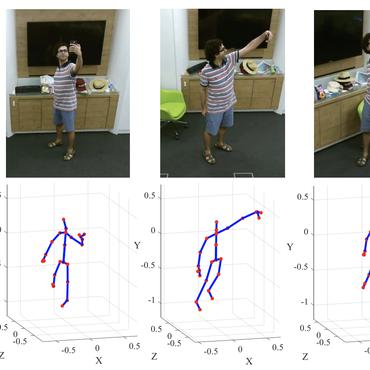}
%
%
\centering
\caption{Skeleton modality as representation for human action}
\label{fig:skeleton}       
\end{figure}

Skeleton data (Fig. \ref{fig:skeleton}) is often obtained through depth sensors or other motion-capturing technologies that track the positions of joints in three-dimensional space. Each joint, such as the head, shoulders, elbows, hips, knees, and ankles, is represented as a point in the skeleton. The connections between these points form a skeletal structure, creating a dynamic representation of human movement.

The integration of vision-based data, encompassing diverse modalities, empowers applications with a holistic and intricate understanding of visual information. This comprehensive approach not only enhances the accuracy and effectiveness of tasks such as fall detection and security but also opens avenues for innovation in the broader field of multimodal data analysis without attracting undue attention.

\subsection{Audio-Based Data}
\label{subsec:3}

Another crucial element of multimodal data sources, particularly for assisted living, is audio-based data \cite{despotovic2022audio}. Voice interactions \cite{xu2019waveear}, ambient sounds \cite{owens2016ambient}\cite{brumm2004impact}, and audio recordings \cite{xu2019waveear} are all included in this category. Audio sensors can be used for sound pattern analysis, voice recognition, and acoustic event detection. They are essential in determining potential emergencies, evaluating residents' well-being, and providing communication aids for individuals with restricted mobility. Systems that are aware of their auditory surroundings can identify odd sounds or distress calls, improving people's safety and general quality of life. The analysis of sound patterns goes beyond simple detection, providing insights into the overall acoustic landscape. This analytical capacity plays a pivotal role in assessing the well-being of residents, offering a non-intrusive means of monitoring their activities. Moreover, in cases where individuals may have limited mobility or communication abilities, audio-based data serves as a cornerstone for developing communication solutions.

By comprehensively understanding the auditory environment, systems can be trained to detect unusual noises or calls for help. This heightened level of situational awareness enhances the safety of residents, enabling timely responses to emergencies and contributing to an improved overall quality of life. The synergy of audio-based data with other multimodal sources further enriches the depth and context of information available to the system, fostering a more holistic understanding of the environment and the individuals within it. In essence, audio-based data plays a pivotal role in creating intelligent and responsive systems that prioritize safety, well-being, and effective communication in various contexts.

\subsection{Sensor Data}
\label{subsec:4}
Sensor data offers a multitude of information about the physical environment and the activities within it, in addition to vision and audio. Wearable devices, motion sensors, temperature sensors, and other types of sensors are included in this category \cite{homayounfar2020wearable} \cite{cheng2021recent}. Sensor data helps with tracking daily activities, keeping an eye on health metrics, and guaranteeing residents' comfort. When deviations from the norm are noticed, the integration of sensor data with other modalities can be used to trigger alerts or responses, providing a more comprehensive picture of an individual's well-being. Motion sensors make a contribution by spotting movement in the surroundings and revealing details about the workings of everyday life. These sensors come in handy for monitoring habits, seeing trends, and deciphering behavioral shifts that might be a sign of health problems or other issues. Contrarily, temperature sensors provide information about a space's thermal conditions, which can be vital for guaranteeing the comfort and wellbeing of its occupants.

Wearable devices, often equipped with a variety of sensors, offer a personalized and continuous stream of health metrics \cite{haghi2021wearable}. These metrics can include vital signs such as heart rate, activity levels, and sleep patterns. The integration of wearable data into the multimodal framework adds a layer of granularity to health monitoring, allowing for a more comprehensive assessment of an individual's physical well-being.

Combining sensor data with visual and auditory information improves our understanding of a person's overall health. Systems are able to produce a more comprehensive and nuanced picture of an individual's health, activities, and environment by combining data from various modalities. Integration of sensor data also makes it possible to create intelligent systems that can react or issue alerts in the event that they detect departures from predetermined norms. By taking a proactive stance, prompt interventions can be made, enhancing resident safety and care quality.

To summarize, the integration of sensor data into multimodal systems enhances their ability to track, evaluate, and react to the intricacies of the physical world and the activities that occur there, promoting a more thorough and individualised approach to wellbeing.

\begin{figure}[h]
\centering
\includegraphics[scale=.3]{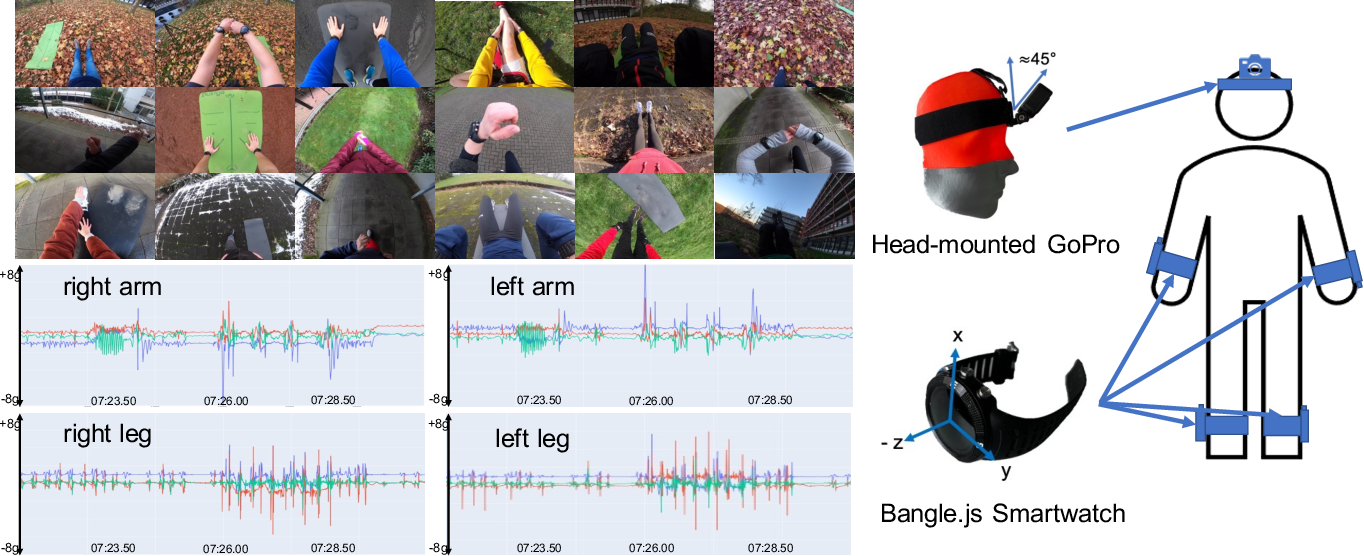}
%
%
\centering
\caption{Sensor output produced by head-mounted camera and smartwatch}
\label{fig:sensor}       
\end{figure}

The sensor outputs generated by a smartwatch and a head-mounted camera (Fig. \ref{fig:sensor}) correspond to different modalities, each of which captures particular physiological signals and elements of the user's surroundings. Visual information is produced by the head-mounted camera, which presents a series of pictures that give the user a firsthand look at their surroundings. This modality is especially useful for lifelogging, augmented reality, and virtual reality applications since it records a wide range of rich and detailed visual details about the surroundings, objects, and activities from the user's perspective. The smartwatch sensor output, on the other hand, is more concerned with physiological signals and tracks data like heart rate, steps taken, and sleep patterns. This data provides insights into the user's health and activity levels, making it suitable for applications related to fitness tracking, health monitoring, and overall well-being. The combination of these modalities contributes to a holistic understanding of both the user's visual experiences and physiological state.

\subsection{Text Data}
\label{subsec:5}

In the context of multimodal data, text data is an additional useful dimension that supplements vision, audio, and sensor inputs. Textual data analysis is the focus of this category, and the data may come from written documents, electronic health records (\cite{ Hayrinen2008definition}), communication logs (\cite{wang2015automatic}), or other sources.

Textual data sheds light on communication patterns, emotional states, and cognitive aspects. By using Natural Language Processing (NLP) techniques, one can understand sentiments, intentions, and contextual nuances by extracting meaningful information from text. This is especially pertinent to applications where it's critical to track mental health or evaluate people's emotional wellbeing.

In healthcare settings, for example, the analysis of text data from medical records can contribute to a more comprehensive understanding of a person's health history, treatment plans, and medication adherence. In a broader context, communication logs or written records can offer valuable information about social interactions, lifestyle preferences, and potential changes in behavior.

Contextual understanding of a person's situation is improved when text data is integrated with other modalities. Multimodal systems obtain a more comprehensive understanding of residents' general well-being by merging textual insights with visual, auditory, and sensor data. A more tailored and flexible response to each person's needs is made possible by this integrated approach, which also makes proactive interventions based on a thorough comprehension of the context possible.

Furthermore, the analysis of text data can contribute to refining the responsiveness of systems, enabling them to generate more natural and context-aware interactions in applications like virtual assistants or communication interfaces.

In conclusion, the inclusion of text data in multimodal systems enriches the analytical depth and contextual understanding, fostering a more comprehensive and nuanced approach to monitoring, care, and communication within various domains.

\subsection{Large Language Model}
\label{subsec:5}

LLMs such as GPT-4 and Flamingo significantly enhance multimodal action recognition systems through advanced contextual reasoning and semantic interpretation. Leveraging extensive world knowledge and linguistic priors, LLMs analyze actions not merely as patterns in raw signals but as meaningful human behaviors situated within a broader context. For example, GPT-4 Vision demonstrates the ability to generate rich textual descriptions of observed actions, and in zero-shot evaluation it surpasses conventional vision models on video benchmarks such as UCF-101 and HMDB-51 \cite{wu2023gpt4v}. This suggests that LLM-based contextual reasoning can effectively handle complex temporal dependencies and ambiguous visual cues.

Flamingo reinforces this capability by integrating vision and text via cross-attention, enabling the language model to attend to visual evidence and infer ongoing activities \cite{alayrac2022flamingo}. With such architectures, LLMs can serve as adaptive fusion controllers: dynamically reweighting modality contributions based on contextual relevance. For instance, an LLM may prioritize audio cues over visual cues in scenes with motion blur, or combine sensor-derived movement features with linguistic priors to infer a more accurate description of an action. Consequently, LLM-driven fusion provides robustness in challenging real-world scenarios, enabling the system to not only classify actions but also reason about intent, environment, and situational context.

\subsection{Integration of Multimodal Data}
\label{subsec:6}

Human activity recognition has benefited greatly from earlier research based on still images. One such study was conducted by Simonyan et al. \cite{simonyan2014two}, who suggested a very deep network for image recognition \cite{simonyan2014very}. This approach utilized two streams of CNN that combine motion and spatial streams and fused them using straightforward averaging and SVM fusion was proposed as a breakthrough. Additionally, it obtains supplementary data, which enhances accuracy. Using transfer learning from the extensive ImageNet dataset, this method inherits the features of image classification for the identification of actions in videos. To fuse or integrate all streams, their suggested technique was expanded to include gated CNNs \cite{yudistira2017gated}.

While each of these multimodal data sources provides valuable insights individually, their true power is realized through integration. The fusion of vision, audio, and sensor data allows systems to paint a more complete picture of the living environment and the needs of the residents. This comprehensive understanding serves as the foundation for adaptive decision-making and personalized assistance, ensuring that active and assisted living systems are not just reactive but proactive in catering to the unique requirements of each individual.

As we move forward in this section, we will explore how deep neural networks can be leveraged to process and fuse these diverse data sources effectively, highlighting the importance of adaptability to the specific preferences and necessities of the residents in active and assisted living environments.

%
%

\subsection{Gating Networks: A Conceptual Overview}

Gating networks, a fundamental element of adaptive fusion, act as controllers for the information flow from different multimodal data sources \cite{kurita2003viewpoint}\cite{o2021and}\cite{yudistira2017gated}. These networks learn to assign different weights to each modality dynamically. Essentially, they determine which data sources are more influential at any given moment, based on the context and the resident's specific requirements. This dynamic modulation of data integration is what makes gating networks particularly well-suited for creating adaptive systems.
The dynamic modulation of data integration through gating networks makes them well-suited for creating adaptive systems. An adaptive system can adjust its behavior based on changing conditions. In this context, the adaptability of gating networks allows for an intelligent and context-aware integration of information from different sources, enhancing the overall performance and responsiveness of the system.

In summary, gating networks play a pivotal role in adaptive fusion by dynamically controlling the flow of information from multimodal sources, assigning varying weights based on context, and thereby creating systems that can intelligently adapt to different situations and user need

\subsection{Adaptive Fusion of Multimodal Deep Networks for Active and Assisted Living}
\label{subsec:5}

With the addition of gating networks, the adaptive fusion of multimodal deep networks for active and assisted living offers a novel way to improve the performance of systems intended for active and assisted living scenarios. In order to facilitate more context-aware decision-making in support of people's daily activities, this research addresses the difficulty of efficiently combining information from various modalities, such as sensor data, audio inputs, and visual cues. In this particular context, the introduction of gating networks has a particularly significant impact because it enables the model to dynamically modify the emphasis on various modalities according to the unique requirements and preferences of users. By adapting the fusion process in real-time through gating mechanisms, the model can provide personalized and responsive assistance, contributing to improved user experiences and overall effectiveness in active and assisted living environments.

In terms of contribution, this work addresses a crucial aspect of multimodal systems for active and assisted living by focusing on adaptability through gating networks. Many existing systems may struggle to account for the dynamic nature of user preferences, health conditions, and environmental factors. The adaptive fusion approach contributes to more resilient and user-centric solutions, offering a level of customization that aligns with the individualized nature of active and assisted living scenarios. The study likely provides evidence, through empirical evaluations, demonstrating the effectiveness of the proposed adaptive fusion method in enhancing the performance and user satisfaction compared to traditional fixed-weight fusion strategies.

In the future, this research may focus on improving and expanding the adaptive fusion technique for more widespread use in assisted and active living. A more thorough grasp of the context might be obtained by looking into the effects of additional modalities and integrating them into the fusion process, such as user feedback or ambient environmental data. The incorporation of explainable AI techniques to improve interpretability and transparency—two important factors in fostering trust with end users and caregivers—may also be explored in future studies. Additionally, deploying the adaptive fusion system in real-world settings, possibly through collaborations with assisted living facilities or home environments, could offer valuable insights into its practicality, scalability, and potential challenges. Overall, this work serves as a significant step toward advancing multimodal systems for active and assisted living and lays the foundation for future research endeavors in this evolving domain.

\begin{figure}[H]
\centering
\includegraphics[scale=.5]{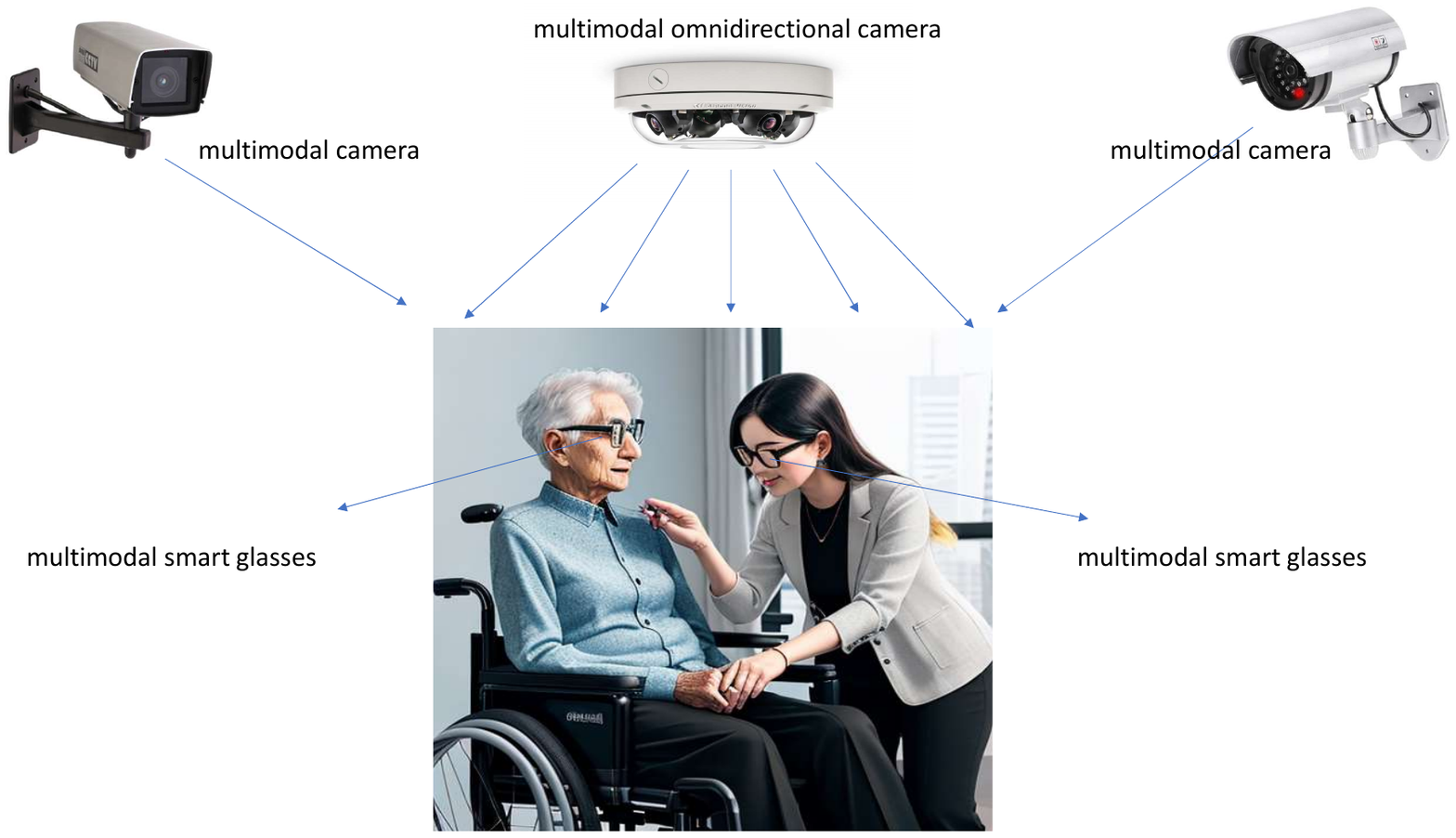}
%
%
\centering
\caption{Implementation of active and assisted living for individuals with special needs utilizes computer vision, where a multimodal camera monitors individuals and their surroundings. Smart egocentric glasses are used to gather information and recommendations about the environment, and this information can be transmitted to security operators.}
\label{fig:example}       
\end{figure}

The implementation of active and assisted living for individuals with special needs incorporates cutting-edge technology, particularly computer vision. In this setup (Fig. \ref{fig:example}), a multimodal camera is deployed to systematically monitor both the individuals with special needs and their surroundings. This camera, equipped with various sensing capabilities, enhances the system's ability to capture a comprehensive view of the environment. To further augment this monitoring process, individuals wear smart egocentric glasses. These glasses serve as a personalized device, providing real-time information and recommendations about the surrounding environment directly to the users. This innovative approach not only caters to the specific needs of individuals but also fosters a greater sense of autonomy and independence. Additionally, the collected information from the egocentric glasses can be efficiently transmitted to security operators. This seamless communication channel ensures that security personnel are promptly informed about any potential issues or emergencies, enabling them to respond swiftly and effectively. Overall, the implementation of active and assisted living, combining computer vision and wearable technology, represents a significant advancement in enhancing the well-being and safety of individuals with special needs.

\section{Methodology}
\label{sec:3}
Deep neural networks have become a game-changer in the quest to build intelligent systems that can comprehend and adapt to the specific needs of people living in active and assisted living environments. These networks are exceptionally well-suited for processing complex data and extracting meaningful insights because they are inspired by the structure and function of the human brain. We examine the use of deep neural networks in assisted and active living in this subsection, highlighting the critical role these networks play in augmenting the capabilities of these systems.

\subsection{Multi-stream Deep Learning Models}

A subset of machine learning called deep neural networks has shown to be an effective tool in a number of fields, such as reinforcement learning, computer vision, and natural language processing. These models can automatically find patterns and features in large and complex datasets because of their capacity to learn hierarchical representations of data. This capability is used to interpret multimodal data obtained from vision sensors, audio inputs, and other sensors in active and assisted living settings.
Deep neural networks' innate hierarchical representation learning is especially helpful in handling the diversity and complexity of multimodal data. By automatically extracting and discerning relevant features from vision, audio, sensor, and text data, these models contribute to a more nuanced and comprehensive understanding of the environment and the individuals within it.

\subsection{Adaptive Fusion Techniques}
\label{sec:4}
In the domain of dynamic and supported living, seamless integration of diverse data streams holds crucial importance in crafting systems attuned to contextual awareness. A key strategy in this pursuit involves employing adaptive fusion methods, with gating networks emerging as a notably efficacious avenue. As a subset within deep neural networks, gating networks inject flexibility into the fusion procedure, enabling systems to customize reactions and engagements according to the distinct requirements and inclinations of inhabitants. Within this section, we delve into the notion of adaptive fusion employing gating networks, underscoring their importance in amplifying the functionalities of systems tailored for active and assisted living settings.

\begin{figure}[H]
\centering
\includegraphics[scale=.4]{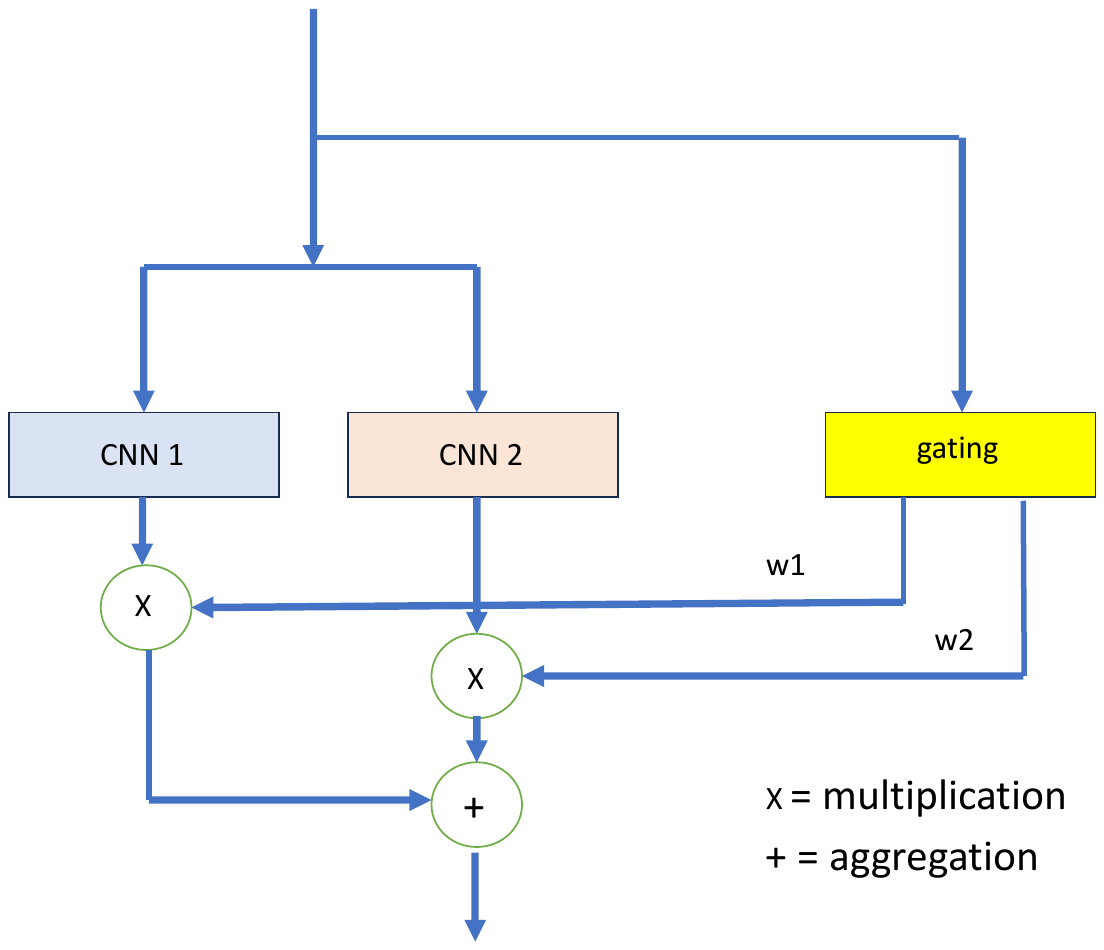}
%
%
\centering
\caption{General gating network architecture which combines CNN 1 and CNN 2 backbones.}
\label{fig:gating1}       
\end{figure}

The general gating network architecture (Fig. \ref{fig:gating1}) represents an advanced framework that combines the features of two Convolutional Neural Network (CNN) backbones, denoted as CNN 1 and CNN 2. In this integrated structure, a gating mechanism is employed to dynamically modulate the information flow between these two CNN backbones. The purpose of this architecture is to optimize the utilization of information from each backbone based on its relevance to the given task. The gating network serves as a decision-making module, determining the importance of features extracted by CNN 1 and CNN 2 for the specific context. By integrating both CNN backbones, the architecture aims to leverage the strengths of each backbone in handling different aspects or complexities within the data. This fusion approach enhances the overall performance and adaptability of the network, making it more effective in tasks such as image recognition, object detection, or other computer vision applications. The general gating network architecture underscores the significance of dynamic feature integration and adaptability in optimizing deep learning models for diverse and complex visual tasks.

\subsection{Multimodal Data Integration with Gating Networks}

Utilizing gating networks to merge multimodal data encompasses acquiring the optimal method for amalgamating information from diverse origins. These networks dynamically adjust the significance of each modality based on the current context, thereby crafting an individualized and adaptable approach to data fusion. This amalgamation stands as a pivotal element for achieving a holistic comprehension of both the environment and the activities of the resident.

The integration facilitated through gating networks is emphasized as crucial for developing a comprehensive understanding of the environment and the resident's activities. This implies that by effectively blending information from varied sources in a contextually aware manner, the system can construct a more thorough and nuanced depiction of the surroundings and the user's behavior.

\begin{figure}[H]
\centering
\includegraphics[scale=.4]{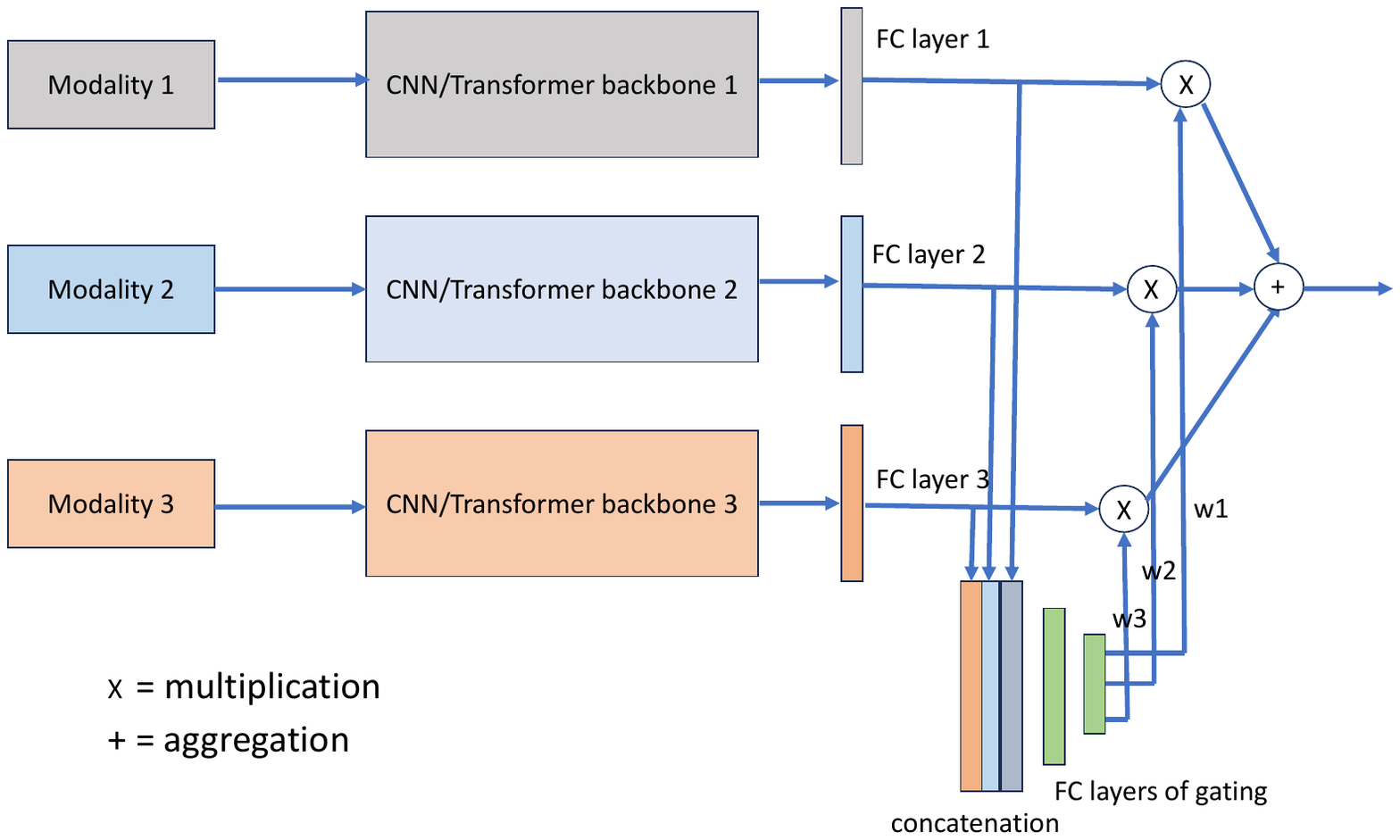}
%
%
\centering
\caption{General gating network architecture which combines 3 modalities network using a multi-layer perceptron of gating network.}
\label{fig:gating2}       
\end{figure}

The General Gating Network Architecture (Fig. \ref{fig:gating2}) is an innovative framework that integrates three distinct modalities within its structure, and this integration is facilitated by a multi-layer perceptron (MLP) serving as the gating network. The three modalities contribute diverse and complementary information, enhancing the network's ability to capture nuanced patterns and features. The role of the gating network, implemented as an MLP, is to dynamically control the flow of information from each modality based on their contextual relevance to the given task. This adaptability allows the network to selectively emphasize or de-emphasize certain modalities, optimizing its performance for specific scenarios. The use of an MLP for gating provides a flexible and non-linear decision-making mechanism, enabling the network to learn intricate relationships and dependencies among the modalities. This architecture is particularly advantageous in tasks requiring the fusion of information from different sources, such as in multimodal recognition or analysis. By combining a multi-layer perceptron gating network with a tri-modal architecture, this approach highlights the importance of adaptive and context-aware feature integration for improved performance in complex tasks that involve multiple types of data.

In summary, the process involves gating networks learning how to optimally combine multimodal data by dynamically adjusting the importance of each modality. This personalized and adaptive approach is crucial for gaining a comprehensive understanding of the environment and the activities of the resident or user.

The weights from the gating model are subjected to the softmax activation function in Equation~\ref{equation2}. The softmax activation function involves dividing the exponential value of one of the outputs (${e^{w_i - \max(\mathbf{w})}}$) by the exponential value of all available outputs(${\sum_{j=1}^{n} e^{w_j - \max(\mathbf{w})}}$). Each weight will be assigned to the softmax activation function:
\begin{equation}\label{equation2}
    z_i = \frac{e^{w_i - \max(\mathbf{w})}}{\sum_{j=1}^{n} e^{w_j - \max(\mathbf{w})}}
\end{equation}

The effect of the softmax activation function is that the total value of the assigned weights will always equal 1. The weights after softmax normalization meet the following condition:

\begin{equation}\label{equation3}
    \sum_{j=1}^{n}z_j = 1
\end{equation}

The final function that determines the final classification ($y$) in Equation~\ref{equation1} is an aggregation of outputs from multiple expert models. The output of each available expert model ($x_n$) is multiplied by the weights provided by the gating model ($z_n$). 

\begin{equation}\label{equation1}
    y = \sum_{j=1}^{n} z_j\bm{x_j}
\end{equation}

\section{Results}
\subsection{RGB and Optical Flow Stream Action Recognition case}
\label{sec:5}
In the field of computer vision, RGB and Optical Flow Stream Action Recognition are two important techniques, especially when it comes to comprehending and analyzing human actions in videos. RGB Action Recognition bases its analysis on the color information that appears in successive video frames. The visual appearance of the scene is captured in each frame, which is represented as a matrix of pixel values in the RGB color space. This technique recognizes and categorizes various actions based on variations in color and intensity over time. It works well for tasks like gesture recognition or sports analysis where understanding an action depends on the motion and appearance of objects.

\begin{figure}[H]
\centering
\includegraphics[scale=.4]{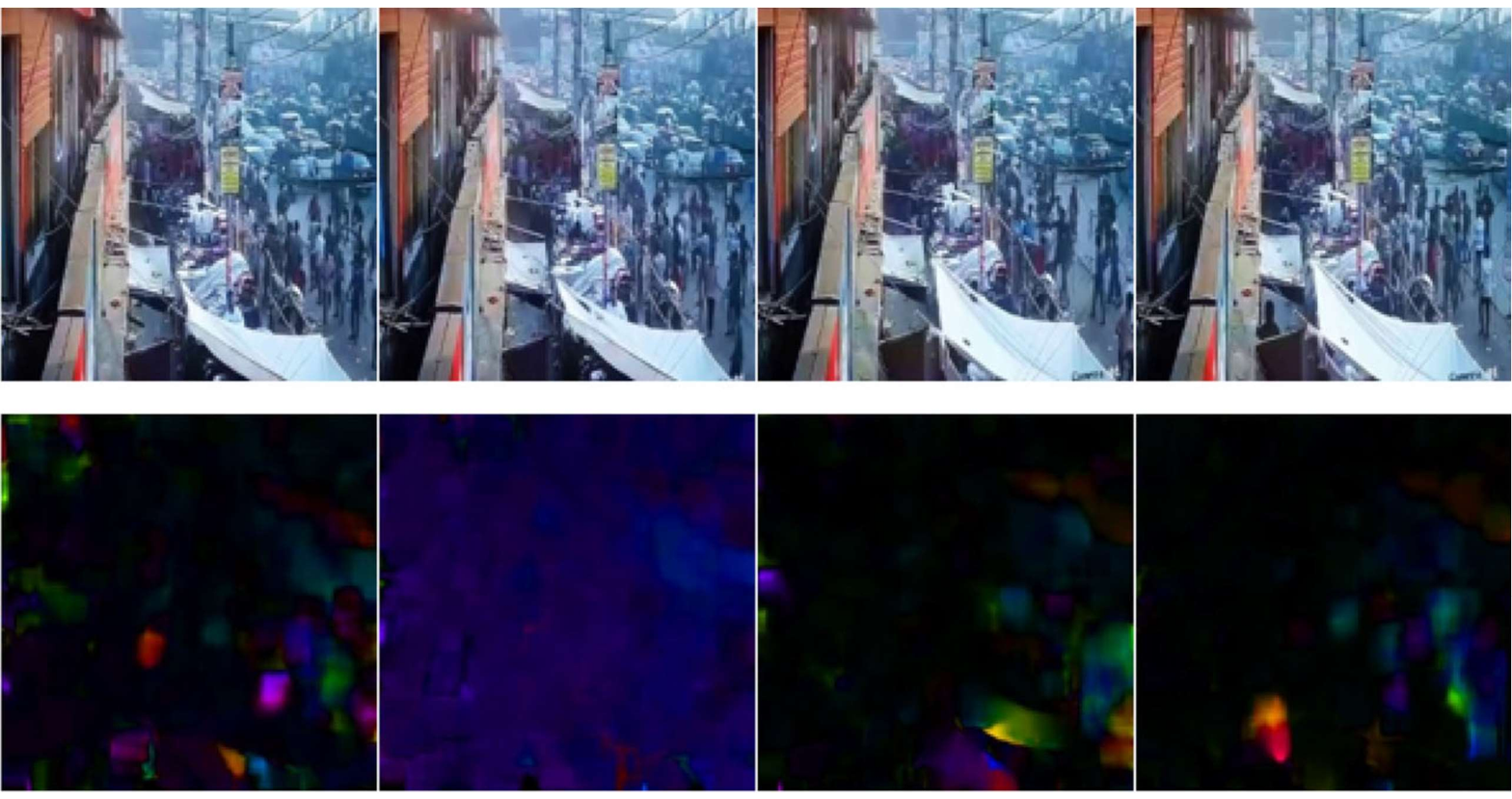}
%
%
\centering
\caption{RGB and Optical flows capture different aspects of features within videos where the former captures spatial information and the latter captures motion information. \cite{pratama2023violence}\cite{cheng2021rwf}}
\label{fig:video}       
\end{figure}

The video in Fig. \ref{fig:video} can be classified using the RGB stream and two-streams, but not with the inclusion of optical flow. This is due to the presence of chaos in a conventional market captured in the video, where numerous objects are recorded by the surveillance camera. Optical flow introduces several biases, such as camera shake or noise, rendering it inaccurate. These biases cause the optical flow to capture the movement of insignificant pixels rather than accurately representing the movement of the objects in the scene.

On the other hand, Optical Flow Stream Action Recognition focuses on the motion of objects by tracking the movement of pixels between consecutive frames. Optical flow algorithms estimate the velocity of objects in an image, providing information about the direction and speed of motion. This approach is particularly useful in scenarios where motion dynamics are essential for action recognition, like in detecting walking, running, or other activities where movement patterns are distinctive. By analyzing the flow of pixels over time, optical flow stream enhances the understanding of temporal aspects of actions, complementing the spatial information provided by RGB analysis. Combining RGB and Optical Flow Stream methods can offer a more comprehensive and accurate representation of complex human actions in video data, providing a robust foundation for action recognition systems.

\begin{figure}[H]
\centering
\includegraphics[scale=.55]{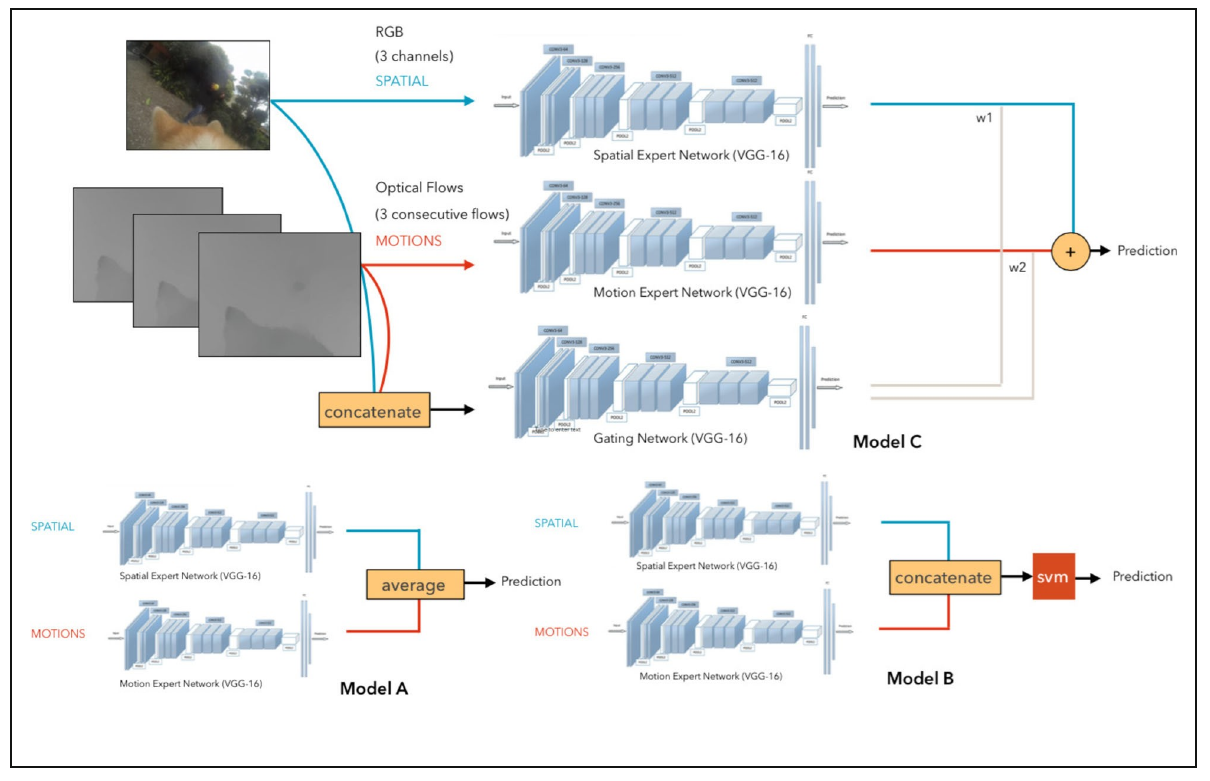}
%
%
\centering
\caption{Model A, model B, and model C are gating network schemes using VGG backbones, ensemble by averaging, and concatenation of two modality networks with SVM, respectively. \cite{yudistira2017gated}}
\label{fig:eurasip}       
\end{figure}

Three different gating network schemes are used for video classification (Fig. \ref{fig:eurasip}), with Model A, Model B, and Model C using VGG backbones as their foundation architecture. Model B uses the process of averaging to adopt an ensemble approach. Creating a final output involves combining the predictions from several VGG-based networks. In contrast, Model A uses a gating network with VGG backbones. This network facilitates an adaptive fusion of information by acting as a mechanism to balance the contributions of various streams or modalities within the ensemble. Lastly, Model C uses VGG backbones to integrate two modality networks through the concatenation technique. This fusion is achieved by concatenating the outputs of the networks and subsequently utilizing a Support Vector Machine (SVM) for classification. Each model represents a unique approach to leveraging VGG architectures for video classification, showcasing the versatility of gating network schemes and ensemble techniques in enhancing model performance.

\begin{table}[H]
  \centering
  \begin{tabular}{ccc}
    \hline
    \textbf{RGB weight} & \textbf{Flow weight} & \textbf{Test accuracy} \\
    \hline
    1.0 & 0.0 & 79.34\% \\
    0.0 & 1.0 & 83.60\% \\
    0.9 & 0.1 & 82.10\% \\
    0.8 & 0.2 & 84.35\% \\
    0.7 & 0.3 & 86.47\% \\
    0.6 & 0.4 & 88.16\% \\
    0.5 & 0.5 & 89.32\% \\
    0.4 & 0.6 & 90.02\% \\
    0.3 & 0.7 & 89.67\% \\
    0.2 & 0.8 & 88.65\% \\
    0.1 & 0.9 & 86.73\% \\
    Gating & & 91\% \\
    \hline
  \end{tabular}
  \caption{Test accuracy for different GB and Flow weights \cite{yudistira2017gated}}
  \label{tab:eurasip}
\end{table}

The test accuracy results for a video classification model under different RGB and Flow weight combinations are shown in table \ref{tab:eurasip}. Each row in the model represents a distinct configuration, indicating the weighting given to RGB and optical flow (Flow). The "RGB weight" and "Flow weight" columns in the model show how much weight is placed on RGB and optical flow, respectively. The resulting accuracy that the model achieves under each configuration is displayed in the third column, "Test accuracy." Notably, a gating mechanism yields the highest accuracy of 91 \%, indicating dynamic weighting according to the video content. As the RGB weight decreases and the Flow weight increases, there is a general trend of improving accuracy, reaching its peak at an even distribution (0.5  \& 0.5) between RGB and optical flow. This table provides valuable insights into the impact of weight distribution on the model's performance, helping to optimize the balance between RGB and optical flow for enhanced accuracy in video classification.

\subsection{RGB and Optical Flow Stream of Violence Detection case}
\label{sec:5}
When it comes to violent behavior detection, RGB and Optical Flow Stream methods are essential for detecting and analyzing visual cues linked to aggression in video footage. Using color information from successive frames, RGB violence detection looks for patterns that indicate aggressive behavior. This means being able to identify changes over time in object interactions, spatial arrangements, and color intensity. For example, sudden shifts in the distribution of colors, quick and erratic movements, or certain body positions can be indicators of aggressive behavior. The RGB approach's efficacy is especially noticeable in situations where scene appearance and contextual details play a major role in the violence recognition task.

Beyond the RGB modality, Optical Flow Stream Violence Detection focuses on the motion dynamics found in video scenes. Optical flow algorithms monitor the motion of pixels between frames to reveal an object's direction and speed. When it comes to identifying violent acts, abrupt and powerful movements—like quick punches or kicks—can be recognized by looking for patterns in optical flow. Optical flow improves the system's ability to identify violent behavior by highlighting the temporal aspects of actions, which is based on the dynamic nature of the scene. By capturing the spatial and temporal nuances inherent in violent behavior, the integration of both RGB and Optical Flow Stream methodologies in violence detection systems enables a more nuanced and robust understanding of aggressive actions. This integrated approach elevates the overall accuracy and reliability of violence detection systems, rendering them valuable tools for applications in security and surveillance.

\begin{table}[h]
    \centering
    \begin{tabular}{lccc}
        \hline
        \textbf{Algorithm} & \textbf{Params (MB)} & \textbf{FPS} & \textbf{Accuracy} \\
        \hline
        Tran et al. \cite{tran2015learning} & 94.8 & 0.5 & 82.75 \\
        Sudhakaran et al. \cite{sudhakaran2017learning} & 94.8 & 0.5 & 77 \\
        Carreira et al.  \cite{carreira2017quo} (RGB only) & 12.3 & 77.3 & 85.75 \\
        Carreira et al.  \cite{carreira2017quo} (OF only) & 12.3 & 77.3 & 75.75 \\
        Carreira et al.  \cite{carreira2017quo} (two-stream) & 24.6 & 59.5 & 81.5 \\
        Cheng et al. \cite{cheng2021rwf} & 0.27 & 94.7 & 87.5 \\
        Ours & 66.6 & 5 & 90.5 \\
        \hline
    \end{tabular}
    \caption{Comparison of algorithm parameters, frames per second (FPS), and accuracy of RWF-2000 dataset. \cite{pratama2023violence}}
    \label{tab:violence}
\end{table}

Three important metrics are used to compare different algorithms in detail: frames per second (FPS), classification accuracy, and algorithm parameters (Params in MB) in the table \ref{tab:violence}. The table's rows each correspond to a particular algorithm, and the columns offer comprehensive details on these important elements. Notable entries are Tran et al. [26] and Sudhakaran et al. [23], which differ in accuracy at 82.75\% and 77\%, respectively, but have the same parameter sizes and frame processing speeds. Different configurations are investigated for Carreira et al. [4], such as RGB-only and optical flow (OF)-only approaches, with a parameter size of 12.3 MB and a consistent frame rate of 77.3. Combining RGB and OF, the two-stream model from Carreira et al. [4] shows a slightly lower FPS of 59.5 and a larger parameter size (24.6 MB), with an accuracy of 81.5\%. The algorithm by Cheng et al. [16] is notable for its low parameters (0.27 MB), high frame rate (94.7), and remarkable accuracy of 87.5\%. Finally, the algorithm we present (Ours) runs at 5 FPS, has a parameter size of 66.6 MB, and achieves a notable accuracy of 90.5\%. This thorough analysis sheds light on the trade-offs between algorithm performance, speed, and complexity in the context of video classification.

\subsection{Mixture of Self-supervised Learning case}

The provided table \ref{tab:gssl} presents classification accuracy results on the Tiny-Imagenet dataset for various augmentation methods applied to a baseline model. The "Method" column enumerates different augmentation techniques incorporated into the baseline, while the "Validation Accuracy (\%)" column quantifies the corresponding accuracy achieved.

\begin{table}[H]
  \centering
  \begin{tabular}{l c}
    \hline
    \textbf{Method} & \textbf{Validation Accuracy (\%)} \\
    \hline
    Baseline & 61.08 \\
    +MixUp & 63.86 \\
    +CutMix & 65.53 \\
    +SmoothMix & 66.65 \\
    +GridMix & 65.14 \\
    +LoRot-E & 66.52 \\
    +LoRot-E + Flip + ChannelPerm & 66.23 \\
    +G-SSL (LoRot-E + Flip) & 66.34 \\
    +G-SSL (LoRot-E + ChannelPerm) & 67.27 \\
    +G-SSL (LoRot-E + Flip + ChannelPerm) & 67.41 \\
    \hline
  \end{tabular}
  \caption{Classification accuracy on the Tiny-Imagenet dataset  \cite{fuadi2023gated}}
  \label{tab:gssl}
\end{table}

The findings highlight the effectiveness of different approaches in the context of self-supervised learning, with a focus on a variety of cutting-edge strategies \cite{fuadi2023gated}. A baseline model has been systematically subjected to various augmentation strategies and self-supervised learning approaches in an effort to improve validation accuracy on the Tiny-Imagenet dataset. The model's performance is gradually improved by adding methods like MixUp, CutMix, SmoothMix, GridMix, and LoRot-E, starting with a baseline accuracy of 61.08\%. Further improvements in accuracy are achieved when these methods are combined in G-SSL (Gated Self-Supervised Learning) configurations involving LoRot-E, Flip, and ChannelPerm augmentations. The highest validation accuracy of 67.41\% is achieved with the comprehensive integration of G-SSL techniques, including LoRot-E, Flip, and ChannelPerm. This table provides a detailed overview of the effectiveness of different self-supervised learning strategies and their synergies in enhancing classification accuracy on the Tiny-Imagenet dataset.

In summary, it demonstrates the effectiveness of augmentation techniques in enhancing the baseline model's performance on the Tiny-Imagenet dataset. The results showcase the incremental impact of individual techniques and highlight the synergies in combining them, particularly within the G-SSL framework, to achieve higher classification accuracies.

\section{Discussion}
\label{sec:5}

\subsection{Real-Time Adaptability Through Gating Networks}

Gating networks play a crucial role in facilitating real-time adaptability in multimodal systems by dynamically adjusting the weights assigned to each modality \cite{yudistira2017gated}. The fusion process is kept adaptable enough to react to new information or shifting circumstances by continuously changing. The flexibility of gating networks is particularly helpful in scenarios such as fall detection, where prompt and context-aware decision-making is crucial. For instance, if something is obstructing the visual data or if there is insufficient lighting, the gating network can swiftly increase the significance of the audio data in assessing the scenario. Owing to its adaptability, the system can determine which information is most crucial and order it appropriately, ensuring dependable performance even under stressful or dynamic circumstances.
 The real-time nature of gating networks contributes to the system's flexibility and responsiveness, making them instrumental in applications where quick and accurate decision-making is essential for user safety and well-being.

Moreover, the continuous weight updating of gating networks promotes resource efficiency and improves adaptability. By dynamically allocating attention to the most informative modalities, the system optimizes computational resources and prevents the processing of less relevant data. This efficiency ensures that the most reliable information source is the focus of computational efforts in the fall detection context, improving the speed and accuracy of decision-making. Because of their resource-conscious operation and real-time adaptability, gating networks are an effective mechanism in multimodal systems that enhance their robustness and effectiveness in various applications such as human-machine interaction, assistive living, and surveillance.

\subsection{Context-Aware Decision-Making with Gating Networks}

Gating networks are critical to enabling systems to make context-aware decisions because they take into account a broader context that encompasses factors like the resident's location, ongoing activity, and immediate needs. With this ability, the system can assess which modalities are appropriate in light of the situation at hand. For example, while preparing a meal, visual data from a camera may be more relevant than audio data when it comes to leisure time. Dynamic adaptation is made possible by gating networks, ensuring that assistance is always customized to the specific context at hand. The ability to make decisions based on context not only improves the system's overall effectiveness but also gives the user a more tailored and responsive experience. This highlights the potential of gating networks in enabling intelligent and adaptive assistive living environments. 

Furthermore, the context-aware decisions made possible by gating networks contribute to an improved user experience by aligning system responses with the resident's immediate needs. This nuanced understanding of context allows the system to deliver timely and relevant assistance, fostering a sense of autonomy and personalized support. In scenarios where residents engage in diverse activities throughout the day, such as cooking, recreational activities, or relaxation, the adaptability of gating networks ensures that the system's interventions are precisely matched to the specific demands of each situation. As a result, residents can benefit from a seamlessly integrated and intelligently responsive assistive living environment that enhances their daily routines and overall well-being.

\subsection{Challenges and Considerations}

Although there are many benefits to using adaptive fusion with gating networks to improve the responsiveness and flexibility of assisted and active living systems, there are drawbacks that must be recognized and resolved. A notable obstacle is the intricacy of gating network training, which necessitates a thorough comprehension of the interactions between various modalities and their contextual significance. Furthermore, biases in the model or training set itself may cause skewed decisions, which would affect the system's accuracy and fairness. It is crucial to take ethical factors into account, particularly when it comes to protecting privacy, to make sure that the gating procedure upholds moral principles and respects individual rights.
A thorough investigation aimed at illuminating these issues will be carried out in this section, with a focus on how critical it is to address these issues in order to ensure the responsible development of assisted and active living systems. Despite these obstacles, gating networks' flexibility and responsiveness are essential to the systems' capacity to meet each person's specific needs, enhancing their autonomy, wellbeing, and quality of life in both assisted and active living environments.

\subsection{Omnidirectional camera based recognition for Active and Assisted Living }

The use of a combination of expert networks can further enrich the personalized and adaptive support that omnidirectional cameras and modality fusion provide in the context of assistive living. A wide range of visual data can be captured by omnidirectional cameras, providing a thorough view of the surroundings \cite{scaramuzza2014omnidirectional}. By combining data from multiple sensors or sources, modalities can be combined to provide a comprehensive picture of the environment. The system can dynamically select and combine the expertise of various networks for particular tasks by integrating a variety of expert networks.

 Because each expert network is tailored to handle specific contexts or aspects of the data, the system can instantly adjust to the varied needs of users. For instance, one expert network might be very good at identifying routine tasks, while another might be especially good at spotting anomalies. By guaranteeing a more thorough examination of the surroundings and enabling prompt responses to people's needs, this collaborative approach maximizes the effectiveness of the assistive living system as a whole, ultimately improving the caliber of care and support offered. The fields of activity recognition and human active assistive living hold great promise for omnidirectional cameras and modality fusion.

\begin{figure}[H]
\centering
\includegraphics[scale=.2]{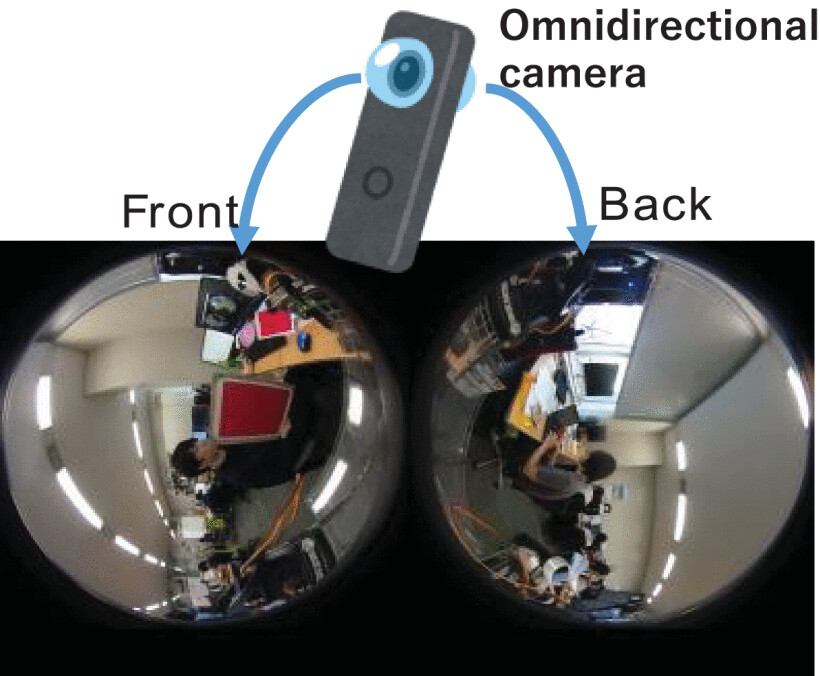}
%
%
\centering
\caption{Omnidirectional camera. \cite{feurstein2018towards}}
\label{fig:omni}       
\end{figure}

\subsubsection{360-Degree Coverage}

With their ability to capture an entire view of the surrounding environment, omnidirectional cameras (as shown in Figure \ref{fig:omni}) are essential for ensuring thorough information retrieval and removing blind spots. This feature is especially helpful for activity recognition systems, where a comprehensive viewpoint is necessary. Omnidirectional cameras provide a more detailed and nuanced image of the surroundings than traditional surveillance, enabling contextual comprehension that is superior. This comprehensive background is essential for correctly identifying and understanding human behavior. The omnidirectional cameras provide 360-degree coverage, which enhances the system's capacity to make decisions based on a comprehensive understanding of the observed context, whether it is monitoring complex movements or understanding interactions within a space.

\subsubsection{Behavioral Analysis and its Privacy Safeguards}

The comprehensive viewpoint that omnidirectional cameras offer not only facilitates detailed examination but also works wonders in understanding complex aspects of human behavior. This ability is essential for spotting trends, spotting abnormalities, and even anticipating possible medical problems or crises based on activity observation. The system's ability to identify minute modifications or departures from standard operating procedures is improved by the comprehensive view, which helps with proactive and preventive health and safety monitoring.

However, as with any technology that involves mass data collection, privacy concerns come first. Strong safeguards are required to guarantee the moral application of collected data because omnidirectional cameras have such a wide field of view. To allay worries and preserve moral principles, privacy precautions like anonymizing personal data and using secure storage techniques must be put into place. Adopting omnidirectional camera systems responsibly in a variety of scenarios requires finding a careful balance between the benefits of in-depth analysis and the privacy of individuals.

A multifaceted approach is needed to address the challenge of privacy concerns in the deployment of omnidirectional camera systems. It involves strong legal and regulatory frameworks in addition to technological solutions. To foster public confidence and guarantee the responsible use of this technology, it is imperative to create and implement explicit policies pertaining to data collection, storage, and utilization. As omnidirectional cameras' capabilities develop further, preventative measures for privacy issues need to change with them.

Omnidirectional cameras offer a multifaceted perspective that is highly valuable for comprehending human behavior and for monitoring health and safety. Notwithstanding, managing privacy apprehensions presents a significant obstacle, necessitating a cautious equilibrium between the advantages of meticulous examination and moral considerations for conscientious implementation in diverse settings.

\subsubsection{Real-Time Responsiveness and Privacy Concern}

The combination of various modalities—particularly the addition of omnidirectional cameras—enables systems to react quickly to detected activity, which is an essential feature in situations where quick assistance or intervention is required. Enhancing the precision of activity identification necessitates the smooth integration of AI and machine learning algorithms for analyzing the combined data. In addition to continuously improving accuracy, these algorithms also dynamically adjust to shifting conditions by taking note of user behavior and changing to meet new demands.

Additionally, omnidirectional cameras and modality fusion are seamlessly integrated into smart home systems, going beyond their immediate applications. This seamless integration has significant implications for the creation of intelligent environments, providing tailored and context-aware support that improves people's quality of life in general. The potential for developing intelligent, adaptable systems creates new opportunities for applications in assisted living, smart homes, and healthcare.

However, the implementation of these cutting-edge technologies presents certain difficulties that need to be carefully considered. Privacy concerns are one significant challenge. The extensive use of omnidirectional cameras in public areas and smart homes poses important concerns about possible privacy violations. The careful development and application of these technologies requires striking a delicate balance between the benefits of increased security and the protection of personal privacy.

In conclusion, it is impossible to overestimate the vast potential of omnidirectional cameras and modality fusion for activity recognition and human assistive living. These technologies make it easier to develop complex, context-aware systems that benefit a variety of industries. To ensure responsible and courteous use across various domains, it is crucial to address privacy concerns and navigate ethical considerations during the development and deployment of such advanced technologies.

\subsubsection{Person Detection using Omnidirectional Camera}
 Person detection and tracking are the main uses of omnidirectional cameras \cite{kobilarov2006people}\cite{boult2004omni}\cite{benli2017human}. However, there are a number of difficulties when integrating deep learning methods—more especially, CNN-based object detectors—for the purpose of person detection in omnidirectional images. First of all, standing people are oriented differently in these photos; instead of facing upward as is typically the case with images taken with side-mounted cameras, they are aligned with the radial axis. Secondly, the equidistant projection of fisheye cameras causes significant object deformation, which restricts the use of pre-trained models and lowers the effectiveness of transfer learning. Furthermore, the unusual look of someone standing directly under the camera presents a situation that is uncommon in conventional perspective photos.

It is crucial to highlight that the absence of a standardized large-scale dataset has compelled researchers to rely on various datasets and metrics for evaluation. Consequently, making direct performance comparisons across studies becomes challenging, underscoring the necessity for standardized evaluation criteria in the realm of omnidirectional camera-based person detection and tracking.

The deployment of CNN-based object detectors for person detection in omnidirectional images presents a number of challenges, including the distortions caused by fisheye cameras and the varying orientation of individuals. Improving the precision and dependability of person detection in omnidirectional camera systems requires the development of strong algorithms that can take these particular difficulties into consideration. To further promote uniformity and comparability amongst research projects in this area, addressing the absence of a shared large-scale dataset and developing standardized evaluation criteria are essential challenges.

Although omnidirectional cameras are a great help for tracking and identifying people, using deep learning techniques in this situation presents some significant challenges. The development of dependable and efficient person detection algorithms in omnidirectional camera systems is imperative due to orientation variations, fisheye camera deformations, and the lack of standard evaluation criteria.

\subsubsection{Object detection}

There has not been much research done on object detection in omnidirectional images, in part because person detection is thought to be more useful and there isn't as much training data available. The development of the THEODORE dataset by Scheck et al. \cite{scheck2020learning} is a notable attempt to close this gap. This dataset includes five classes of people with accompanying them: armchair, chair, table, TV, and wheeled walker. THEODORE was used to train a variety of object detection models, such as SSD, R-FCN, and Faster R-CNN. The FES dataset was then used for testing. The CenterNet model performed better in omnidirectional object detection when unsupervised domain adaptation (UDA) techniques were incorporated, namely entropy minimization (EM) and maximum squares loss (MSL).
Due to its limited use cases, object detection in omnidirectional views has not been used much, but it has great potential for complex tasks like action recognition in smart monitoring systems, especially in the elderly care industry. The lack of training data and the idea that object detection in omnidirectional images is less useful than person detection remain major obstacles in this field. To fully utilize object detection in omnidirectional views, especially for applications in smart monitoring systems, these issues must be resolved.

The overhead fisheye camera has shown to be useful in human pose estimation (HPE) for both 2D and 3D pose estimation. Instead of estimating joint positions, Georgakopoulos et al. \cite{georgakopoulos2018pose} trained a CNN to distinguish pre-set postures using a 3D human model. For joint position estimation, Denecke and Jauch \cite{denecke2021verification} used a smart sensor's 3D point cloud and their prior understanding of the human body. Rectilinear views produced from omnidirectional images were utilized by Heindl et al. \cite{heindl2019large} for OpenPose-based pose estimation. Garau et al. \cite{garau2021panoptop} accomplished viewpoint-invariant 3D pose estimation in a top-view scenario without explicitly using omnidirectional images. The applications of HPE extend to Human Activity Recognition (HAR), contributing to the development of systems for monitoring activities, such as elder care.

There are many different obstacles and possible uses for the investigation of object detection and human pose estimation in omnidirectional images. The development of computer vision in smart monitoring systems is influenced by efforts in model training, dataset creation, and novel techniques. This has implications for a number of industries, including elder care.

\subsubsection{Egocentric 3D pose estimation}

Fisheye cameras are used in a top-view scenario for egocentric pose estimation, a specialized application that presents unique problems and creative solutions. In this setup, the camera is attached to a device on the subject's head to record the front side of the body as well as the surrounding periphery. EgoCap \cite{rhodin2016egocap} provides an example implementation that uses two fisheye cameras mounted on a bike helmet with a wooden frame that is either Y- or T-shaped. A motion capture system was used to create the training dataset, and ResNet101 was adjusted to create heatmaps for 18 joints. With the help of a user-specific 3D body model and the 2D skeleton, real-time 3D skeleton construction was accomplished.

Some implementations, like xR-EgoPose/SelfPose \cite{tome2019xr} and Mo2Cap2 \cite{xu2019mo}, used single fisheye cameras and produced artificial datasets for training. A CNN with two branches—one for the entire body and another for the lower body—was used by Mo2Cap2 to create heatmaps. Projecting joints into 3D coordinates was made easier by estimating the distance between the joints and the camera. xR-EgoPose combined a lifting module for 3D pose regression and high-resolution 2D heatmap output with ResNet101 for joint heatmap generation. Using picture sequences and motion priors from the AMASS dataset, Wang et al. presented a method for global pose estimation that estimates 3D joint positions in the world coordinate system. They mounted a single fisheye camera on a helmet as part of their setup.

EgoGlass \cite{zhao2021egoglass} is an example of an innovative method; it mounts two cameras on regular eyeglasses to reduce the size of the apparatus and captures both sides of the body for pose estimation in stitched-together images. These developments highlight the variety of egocentric pose estimation applications and approaches, as well as the adaptability of fisheye cameras in obtaining distinctive views for human pose analysis.

The variety of applications for egocentric pose estimation demonstrates how fisheye cameras can be used to capture unusual angles for human pose analysis. The use of artificial datasets, creative setups, and both single and dual fisheye cameras highlights the diverse nature of this field of study.

Developing techniques to handle differences in camera placement, device size, and capturing angles is a major challenge in egocentric pose estimation. Attaining uniformity in pose estimation amongst various configurations is crucial to promote wider relevance and comparability of outcomes in this field.

\subsubsection{Action recognition}

Using top-view fisheye camera images, a novel method for action recognition was put forth \cite{wang2019mask}. In order to dewarp the spherical image into a panoramic view, the technique used Mask-RCNN \cite{he2017mask} to identify the spine lines of standing individuals. Mask-RCNN was used to perform person detection on the panoramic image, and max pooling of bounding boxes across frames was used to obtain Regions of Interest (ROIs). In order to minimize computational costs, a binary mask derived from the ROIs was used for action recognition on the 16 frames using a 3D ResNet. A network that estimates scores for different actions was trained using multi-instance multi-label learning (MIML) \cite{zhou2012multi}. Stephen et al. expanded on this strategy by employing RAPiD for person detection to create stacked feature maps for every person in the omnidirectional image and by introducing a parallel pipeline for people in the central area \cite{stephen2021hybrid}. Because of their larger field of view (FOV), omnidirectional images have been studied in more depth than the previously mentioned highly researched topics \cite{yu2023applications}.

The ODIN dataset \cite{ravi2023odin}, which provides full-body 3D pose estimates from top-view omnidirectional images, is a crucial advancement in this field. This dataset is expected to be expanded to include a wider range of data, as it is proving to be extremely useful for training models in omnidirectional 3D pose estimation. These developments demonstrate how constantly changing the field of human behavior understanding is, as they stand to significantly aid in overcoming a variety of challenges.

\begin{table}[h]
    \centering
    \begin{tabular}{l r}
        \hline
        \textbf{Modality/Characteristic} & \textbf{Amounts} \\
        \hline
        Omnidirectional RGB images & 332K \\
        Lateral-view RGB images & 1.464M \\
        Lateral-view infrared images & 1.464M \\
        Lateral-view depth images & 1.453M \\
        \hline
    \end{tabular}
    \caption{Amounts of different modalities/characteristics in the ODIN dataset.}
    \label{tab:modality_amounts}
\end{table}

The compositional details of the dataset, as shown in Table \ref{tab:modality_amounts}, provide an extensive overview of different modalities and attributes. This dataset contains a variety of image types, all of which have been carefully categorized according to their modality. The data type is accurately identified by the modality/characteristic column, which includes Omnidirectional RGB images, Lateral-view RGB images, Lateral-view infrared images, and Lateral-view depth images. The corresponding amounts column counts the number of occurrences for each modality. For example, Omnidirectional RGB images have 332,000 instances, while Lateral-view RGB and Lateral-view infrared images have roughly 1.464 million instances.
 1.453 million lateral-view depth images, with a slight deviation. This brief table offers a useful overview of the structure of the dataset and sheds light on the distribution of various data modalities and their corresponding quantities. When navigating the dataset, this information is invaluable for researchers and practitioners as it allows for a more nuanced understanding of its diversity and possible uses.

The use of omnidirectional cameras in Human Behavior Understanding (HBU) represents an emerging field, with a primary focus on addressing pose classification challenges. Despite the scarcity of datasets specifically designed for this purpose, efforts have been made to leverage omnidirectional cameras for HBU applications. One notable approach proposed using multiple omnidirectional cameras to detect human activities in indoor environments. This novel approach simplifies participant tracking and categorizes four common indoor activity postures.
 Another noteworthy contribution involves the formulation of a pose classification method that utilizes silhouettes captured by omnidirectional cameras, sorting them into three distinct poses: falling, sitting, and standing.

The scarcity of specialized datasets presents a significant challenge when integrating omnidirectional cameras for Human Behavior Understanding. Establishing dedicated datasets tailored for specific HBU applications, as well as addressing issues of consistency and standardization across various scenarios, is critical for advancing research in this rapidly evolving field.

\subsection{Increased Accuracy of Multimodal Fusion}

Multimodal fusion techniques are critical for improving the accuracy of activity recognition systems because they seamlessly integrate information from multiple sources, including visual, audio, and sensor data. This integration takes advantage of the inherent ability of different modalities to provide complementary information, enriching our overall understanding of observed activities. The synergy generated by this fusion process is critical to the system's improved ability to detect intricate details in the observed environment.

One significant advantage of using multimodal fusion is the increased robustness it provides to the system against various types of variability. By combining data from multiple sources, the system becomes more resilient to challenges posed by environmental changes, fluctuations in lighting conditions, and occlusion. This increased robustness is especially important in real-world applications where operating environments are dynamic and unpredictable. It is critical that the activity recognition system performs effectively across a wide range of conditions in order to be practical and relevant in a variety of scenarios.

Looking ahead, adaptive fusion using a gating network appears to be a promising avenue for the evolution of multimodal fusion techniques in activity recognition systems. This novel approach, aided by a gating network, has the potential to dynamically adjust the contribution of each modality based on contextual information, improving the system's adaptability and performance across a variety of scenarios. As we move forward, the incorporation of adaptive fusion via gating networks emerges as a strong contender, promising to improve the robustness and accuracy of activity recognition systems in dynamic and unpredictable real-world environments.

\subsection{Future Directions and Implications}
\label{sec:7}
The trajectory of adaptive fusion for active assisted living with omnidirectional cameras indicates significant future advancements in a variety of fields. Integrating multi-modal sensor data, including technologies such as depth sensors, infrared capabilities, and wearables, is an important avenue for exploration. This comprehensive integration aims to improve the system's adaptability and resilience, resulting in a better understanding of the environment.
 Expected progress in machine learning, particularly in deep learning architectures, is anticipated to play a pivotal role in refining the fusion of omnidirectional camera data, enabling models to autonomously learn and adapt to novel environments and user behaviors for continuous improvement.

Future development will be focused on real-time context awareness, with semantic segmentation and object recognition techniques playing critical roles. This increased awareness will allow the system to dynamically tailor responses based on observed activity, resulting in a more responsive and personalized user experience. Ongoing research may focus on sophisticated human behavior analysis, with the goal of identifying intricate interactions, gestures, and abnormal behaviors. This nuanced understanding is critical for providing personalized assistance that is aligned with users' changing needs and preferences.

Addressing privacy concerns entails combining privacy-preserving techniques with edge computing solutions for localized data processing. Furthermore, using a user-centric design approach, informed by feedback from end users, ensures that the technology aligns with the preferences and values of those it is intended to help. Long-term monitoring and predictive analytics, combined with collaborative research initiatives and interdisciplinary approaches, will drive the evolution of adaptive fusion systems. The desired outcome is a set of innovative solutions that not only improve the well-being and independence of people in assisted living facilities, but also contribute to the larger landscape of technological advancements in healthcare and assistive technologies.

Despite promising results, several opportunities remain for advancing LLM integration in action recognition. A primary challenge is computational efficiency: existing LLMs are large and resource-intensive, motivating research into lightweight architectures, model compression, or knowledge distillation for real-time deployment \cite{liang2024fusion}. Another area of development is \emph{open-vocabulary action recognition}, where the system recognizes unseen actions solely from textual descriptions or minimal examples—leveraging the LLM’s world knowledge without requiring full retraining.

Furthermore, improving multimodal alignment remains essential. Ensuring that sensor, audio, and visual signals are coherently mapped into a unified representation space will be central to achieving fine-grained reasoning over human behavior. Ultimately, deeper LLM integration may yield systems capable not only of classifying actions but also of explaining the underlying context and intent, marking a significant step toward more human-like understanding. Such advancement aligns with broader visions of general-purpose intelligent systems with unified multimodal reasoning capabilities \cite{tandon2025agi}.

\section{Conclusion}
\label{sec:8}
We present a forward-thinking paradigm for improving the quality of life for people in need of care and support. By seamlessly integrating various data sources and employing adaptive deep learning techniques, this study lays the groundwork for more effective and personalized assistance in active and assisted living settings. However, it is critical to recognize that the success of this approach is dependent on effectively addressing ethical concerns and constantly developing fusion methods to meet the changing needs of users in real-world situations. As technology advances, the adaptive fusion framework's potential transformative effects on the well-being and independence of people requiring assisted living services become more promising.

\bibliographystyle{plain}
\bibliography{bibliography}
\end{document}